\documentclass[11pt]{article}

\usepackage[final]{acl}
\makeatletter
\@ifpackageloaded{lineno}{}{}
\makeatother

\usepackage{times}
\usepackage{latexsym}
\usepackage{hyperref}

\usepackage[T1]{fontenc}

\DeclareFontShape{T1}{ptm}{m}{scit}{<->ssub * ptm/m/sc}{}
\DeclareFontShape{T1}{ptm}{bx}{scit}{<->ssub * ptm/bx/sc}{}

\usepackage[utf8]{inputenc}

\usepackage{microtype}

\usepackage{inconsolata}

\usepackage{graphicx}

\usepackage{booktabs}
\usepackage{amsmath}
\usepackage{amssymb}
\usepackage{multirow}
\usepackage{tikz}
\usetikzlibrary{shapes,arrows,positioning,fit,backgrounds}

\usepackage{pifont}
\newcommand{\cmark}{\ding{51}}
\newcommand{\xmark}{\ding{55}}

\newcommand{\system}{Health\textsc{CUES}}

\newcommand{\mllm}{Qwen3-Omni}

\title{From Sound to Symptom: Real-Time Respiratory Signal Understanding for Conversational Healthcare Agents
}

\author{
 \textbf{Tanmay Laud\textsuperscript{1}},
 \textbf{Herprit Mahal\textsuperscript{1}},
 \textbf{Subhabrata Mukherjee\textsuperscript{1}}
\\
 \textsuperscript{1}Hippocratic AI
\\
 \small{
   \textbf{Correspondence:} \href{mailto:tanmay@hippocraticai.com}{tanmay@hippocraticai.com}
 }
}

\begin{document}
\maketitle

\begin{abstract}
Cough events during live spoken conversations carry clinically valuable respiratory signals, yet existing dialogue systems treat them as acoustic noise to be discarded. We present \system{} 
 (Clinical Understanding from Embodied Sounds), a streaming pipeline for paralinguistic respiratory monitoring in real-time conversational agents, a capability that, to the best of our knowledge, is absent from all prior systems. \system{} processes audio through a rolling buffer aligned with dialogue turn boundaries, enabling sub-second event detection without interrupting conversational flow. Beyond binary cough detection, the system provides fine-grained analytics: (i) differentiation between coughing and throat clearing, (ii) cough subtype classification (dry, wet, barking, whooping) with confidence scores, and (iii) temporal duration estimation with start--end boundaries. To prevent alert fatigue, \system{} introduces dialogue-aware gating mechanisms that modulate triggering based on conversational context. The system leverages \mllm{}, a multimodal large language model (MLLM), with constrained structured outputs, decomposing cough analysis into parallel prediction tasks for independent prompt optimization. Evaluation on 847 in-house conversational audio segments demonstrates 93\% F1 for cough detection, 0.75 weighted-F1 for wet/dry subtype classification, and average end-to-end latency of 340ms; external validation on the AMI meeting corpus confirms robust cough, throat-clearing, and speech separation in the presence of speech (0.91 macro-F1). A user study with licensed healthcare professionals confirms the clinical relevance of subtype information and the system's utility in telehealth workflows.
\end{abstract}

\section{Introduction}

Respiratory sounds, particularly coughs, serve as important biomarkers in healthcare conversations. During telehealth consultations and remote health monitoring, cough characteristics provide actionable clinical insights, distinguishing productive from non-productive cough or identifying patterns such as barking or whooping cough \citep{porter2019diagnostic,imran2020ai4covid}. Yet \textbf{current conversational AI systems discard these paralinguistic health signals entirely}, routing them to noise suppression rather than clinical analysis. This gap has concrete consequences: audio-based cough assessment is among the most valuable remote diagnostic tools available, yet no existing conversational agent can leverage it \citep{porter2019diagnostic}.

Building a real-time cough analysis system for live conversations presents distinct challenges: latency constraints require detection within the natural pause between utterances ($<$500ms); telephony codecs and background noise substantially degrade signal quality; clinical utility demands subtype differentiation beyond binary detection; and repeated cough mentions become intrusive without dialogue-aware modulation.

We present \system{} (Clinical Understanding from Embodied Sounds), a streaming cough analysis pipeline that addresses these challenges. Our contributions are:
\begin{enumerate}
    \item A \textbf{streaming cough detection framework} with turn-aligned audio processing for real-time conversational integration (Section~\ref{sec:system}).
    \item \textbf{Fine-grained cough analytics} providing four-way subtype classification, confidence scores, and temporal duration boundaries; to the best of our knowledge, these capabilities are absent from all prior cough detection systems (Section~\ref{sec:system}).
    \item \textbf{Dialogue-aware gating} with cooldown, topic suppression, and cough-to-speech ratio monitoring to prevent alert fatigue (Section~\ref{sec:gating}).
    \item \textbf{Systematic evaluation} with baseline comparisons and licensed healthcare professional validation across three dialogue quality dimensions (Section~\ref{sec:evaluation}).
\end{enumerate}

\section{Related Work}

\paragraph{Audio Event Detection and Cough Analysis.}
General-purpose audio event detection \citep{kong2020panns,gong2021ast} targets broad taxonomies on pre-segmented clips without health-relevant distinctions. Cough analysis spans signal processing \citep{drugman2013objective} and deep learning on dedicated corpora \citep{orlandic2021coughvid,sharma2020coswara}, with COVID-19 spurring interest in cough-based screening \citep{imran2020ai4covid,laguarta2020covid}. Existing systems share three critical limitations: they analyze isolated recordings rather than streaming audio, perform binary detection without subtype differentiation, and lack any notion of dialogue context for appropriate triggering.

\paragraph{Paralinguistic Signals in Dialogue.}
Physiological signals have been used to adapt dialogue behavior \citep{litman2014sigdial,katada2020sigdial}, and video-based respiratory estimation has been integrated into human-robot dialogue to prevent speech collisions and synchronize robot breathing \citep{obi2024sigdial}. Spoken dialogue systems adapting to detected user affect \citep{litman2014sigdial} demonstrate that such signal-to-policy integration improves task success and user satisfaction, an evaluation dimension we adopt. LLM-based affect recognition \citep{zhao2024affect} further demonstrates community appetite for non-lexical signal integration. In contrast to these antecedents, \system{} addresses respiratory health events specifically, providing fine-grained subtype output and dialogue-aware gating designed for clinical triage contexts.

\paragraph{Multimodal Foundation Models.}
Recent MLLMs demonstrate strong zero-shot audio understanding capabilities \citep{gong2023whisper,chu2023qwen,qwen3omni}. \system{} leverages these capabilities, specifically those of \mllm{} \citep{qwen3omni}, while adding structured output constraints and dialogue-aware post-processing essential for reliable real-time operation, additions not addressed by the foundation models themselves.

\section{System Overview}
\label{sec:system}

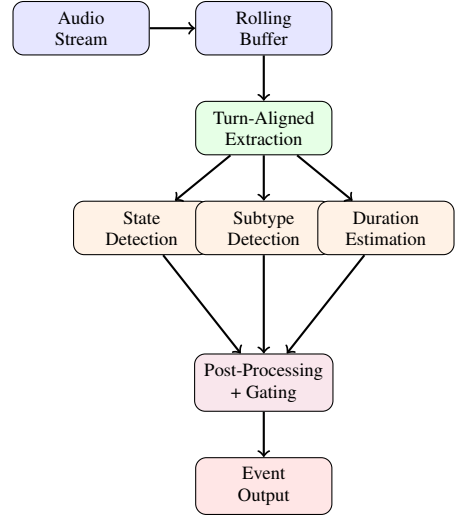
\begin{figure}[t]
    \centering
    \begin{tikzpicture}[
        node distance=0.6cm,
        box/.style={rectangle, draw, rounded corners, minimum width=1.8cm, minimum height=0.6cm, align=center, font=\scriptsize},
        arrow/.style={->, thick}
    ]
        \node[box, fill=blue!10] (audio) {Audio\\Stream};
        \node[box, fill=blue!10, right=of audio] (buffer) {Rolling\\Buffer};

        \node[box, fill=green!10, below=of buffer] (extract) {Turn-Aligned\\Extraction};

        \node[box, fill=orange!10, below left=0.6cm and -0.2cm of extract] (detect) {State\\Detection};
        \node[box, fill=orange!10, below=of extract] (subtype) {Subtype\\Detection};
        \node[box, fill=orange!10, below right=0.6cm and -0.2cm of extract] (duration) {Duration\\Estimation};

        \node[box, fill=purple!10, below=1.3cm of subtype] (post) {Post-Processing\\+ Gating};

        \node[box, fill=red!10, below=of post] (output) {Event\\Output};

        \draw[arrow] (audio) -- (buffer);
        \draw[arrow] (buffer) -- (extract);
        \draw[arrow] (extract) -- (detect);
        \draw[arrow] (extract) -- (subtype);
        \draw[arrow] (extract) -- (duration);
        \draw[arrow] (detect) -- (post);
        \draw[arrow] (subtype) -- (post);
        \draw[arrow] (duration) -- (post);
        \draw[arrow] (post) -- (output);

    \end{tikzpicture}
    \caption{Architecture of \system{}. Audio is buffered and extracted at turn boundaries, then processed by parallel inference tasks. Results are combined with dialogue-aware post-processing before output.}
    \label{fig:architecture}
\end{figure}

Figure~\ref{fig:architecture} illustrates the \system{} architecture, operating as a module within a conversational agent pipeline.

\subsection{Inputs and Streaming Context}

The pipeline receives continuous audio from telephony or microphone input. A \textbf{rolling audio buffer} maintains recent audio history, enabling retrospective analysis when a user turn completes. Extraction is \textbf{turn-aligned}: when ASR indicates end-of-utterance, the system analyzes a window extending slightly before turn onset (default: 2s look-back) to capture cough events that precede speech. Operating at the turn level matches the granularity at which the dialogue agent plans and acts; because the look-back window spans the preceding utterance, coughs that occur mid-turn or just before speech onset are still captured, so the turn-level trigger does not discard out-of-turn respiratory events.

\subsection{Multi-Stage Inference}

Rather than a monolithic classifier, \system{} decomposes cough analysis into parallel structured prediction tasks via \mllm{} \citep{qwen3omni}, a multimodal LLM (MLLM) backend. This decomposition enables independent per-component prompt optimization and graceful degradation if any single task fails.

\textbf{State Detection} performs three-way classification: \textit{coughing}, \textit{throat clearing}, or \textit{none}. Throat clearing is explicitly distinguished, as it carries different clinical significance and warrants different conversational responses.

\textbf{Subtype Classification} categorizes detected cough events into four commonly used clinical descriptors of cough character \citep{morice2007ers,chung2009semantics,porter2019diagnostic}: \textit{dry} (non-productive, no mucus expelled), \textit{wet} (productive, with a rattling or gurgling quality from airway secretions), \textit{barking} (seal-like and low-pitched, classically associated with croup), and \textit{whooping} (paroxysmal fits followed by an inspiratory ``whoop'', associated with pertussis), each with an associated confidence score. These are qualitative descriptors rather than standardized diagnostic categories \citep{chung2009semantics}, which we revisit when reporting subtype agreement (Section~\ref{sec:evaluation}).

\textbf{Duration Estimation} outputs start and end timestamps for each event, enabling severity bucketing (short: $<$2s; medium: 2--6s; long: $>$6s) as a clinical proxy for severity. All outputs follow a constrained schema: subtype returns code-confidence pairs, e.g.\ \texttt{[wet][0.88]}; duration returns time ranges, e.g.\ \texttt{[1.2--4.3]}.

\subsection{Dialogue-Aware Gating}
\label{sec:gating}

A central contribution of \system{} is its gating layer. Without conversational awareness, even accurate detection would be counterproductive: alerting on every cough would disrupt conversation flow, and repeating the same follow-up after each detection would feel robotic. Three mechanisms work together to decide whether a detected event should trigger an agent action.

The pipeline maintains a 60-second sliding window of activity and computes the fraction of recent speaking time occupied by coughing. This \textit{cough-to-speech ratio} ($\text{CSR} = \frac{\sum \text{cough\_dur}}{\sum \text{speech\_dur}}$) lets the system treat a brief isolated cough differently from a sustained pattern that warrants more active follow-up. Once an action has been triggered, a configurable quiet period prevents re-triggering while the patient is still coughing through the same episode. Separately, if the agent's recent turns already contain cough-related language, new detections are suppressed, since the system infers the topic is already being addressed.

These mechanisms interact with the detection outputs: clinically urgent subtypes (wet, barking, whooping) and prolonged episodes clear lower thresholds for action than brief dry coughs.

\subsection{Output and Dialogue Integration}

When a detection clears the gating thresholds, the pipeline produces a natural-language action suggestion passed to the agent's planning module via a lightweight callback API. For example, a 3 second wet cough after two events produces:

\begin{quote}
\small\textit{System action: ``Patient has coughed twice. Sounds like a wet cough. Ask about it if not addressed.''}

\smallskip
\textit{Agent: ``I noticed you've been coughing. Does it feel like you're bringing anything up?''}
\end{quote}

Table~\ref{tab:scenarios} illustrates how subtype and gating interact across three representative scenarios.

\begin{table}[t]
\centering
\footnotesize
\setlength{\tabcolsep}{4pt}
\begin{tabular}{p{0.27\columnwidth}p{0.27\columnwidth}p{0.31\columnwidth}}
\toprule
\textbf{Signal} & \textbf{Gating} & \textbf{Action} \\
\midrule
Dry cough $\times$2, conf.\ 0.79, 1.8s & Count threshold met ($\geq$2 events) & ``Has that cough been bothering you?'' \\
\midrule
Barking cough, conf.\ 0.91, 4.2s & Long duration $\rightarrow$ immediate trigger & ``That cough sounds quite harsh. How long have you had it?'' \\
\midrule
Wet cough, conf.\ 0.84; agent recently mentioned ``cough'' & Topic suppression active & \textit{(suppressed; topic already in dialogue)} \\
\bottomrule
\end{tabular}
\caption{Representative \system{} outputs illustrating subtype-sensitive responses and dialogue-aware gating. Barking and wet coughs trigger at lower thresholds due to higher clinical significance; topic suppression prevents redundant follow-up.}
\label{tab:scenarios}
\end{table}

\section{Evaluation}
\label{sec:evaluation}

\paragraph{Setup.}
We evaluate primarily on 847 held-out conversational audio segments (12.3 hours) spanning telephony-quality recordings (8kHz $\mu$-law) and direct microphone capture (16kHz PCM), with SNR ranging from 5--30dB. These segments were collected in-house to match the conditions of our target telehealth deployment, including its codecs, ambient noise, and overlap between coughs and conversational speech, which public isolated-cough corpora do not reflect. Three U.S.-licensed clinicians labeled event presence, subtype, and timestamps; inter-annotator agreement was substantial for event presence ($\kappa$=0.82) and moderate for subtype ($\kappa$=0.67), reflecting inherent subjectivity. \system{} runs online as a real-time service: it does not require frame-level streaming but instead operates over a rolling audio buffer that is analyzed at turn boundaries during a live call. Our evaluation uses this same rolling-buffer mechanism, without oracle segment boundaries. To probe generalization beyond our own data, we additionally validate on the AMI Meeting Corpus \citep{carletta2006ami} using publicly released cough annotations,\footnote{\url{https://github.com/paulleamy/AMI-Cough-Annotations}} which contain spontaneous coughs, throat clears, and speech recorded in multi-party meetings, a demanding in-the-presence-of-speech setting.

\paragraph{Results.}
Table~\ref{tab:results} shows \system{} achieves 93\% F1 for cough detection, 0.75 weighted-F1 on the clinically central wet-versus-dry subtype distinction, and 340ms average end-to-end latency, well within the 500ms conversational pause budget. We include generic audio encoders (BEATs \citep{chen2023beats}: 59.1\% F1) and cough-specialized models (PANNs \citep{kong2020panns}: 76.4\%; CoughVID fine-tuned \citep{orlandic2021coughvid}: 79.8\%) as reference points. These baselines are \emph{not} compute- or backbone-matched: they differ from \system{} in model scale, pretraining data, and task formulation, so the comparison is capability-illustrative rather than a controlled head-to-head, and should not be read as a like-for-like accuracy ranking. Its purpose is to situate \system{} against the off-the-shelf detectors a practitioner might otherwise adopt, none of which provide the streaming, throat-clearing separation, subtype, duration, and dialogue-aware capabilities that \system{} unlocks from a single general-purpose MLLM. The CoughVID fine-tuned model attains the lowest latency (95ms) but is restricted to binary detection; \system{}'s higher latency (340ms) reflects its richer, multi-task inference. Table~\ref{tab:capabilities} confirms \system{} is the only evaluated system providing all five capabilities essential for clinical dialogue use.

\paragraph{External Validation (AMI).}
To test whether detection holds beyond our in-house recordings, we evaluate on the AMI Meeting Corpus \citep{carletta2006ami}, where coughs and throat clears co-occur with multi-party conversational speech. With no adaptation to that domain, \system{} sustains strong three-way performance: macro-F1 0.91 and accuracy 0.92 ($n{=}719$), with per-class F1 of 0.86 (coughing), 0.94 (throat clearing), and 0.92 (none); see Table~\ref{tab:ami}. 

\paragraph{Subtype Breakdown.}
The clinically central distinction is wet versus dry (productive vs.\ non-productive), which reaches 0.75 weighted-F1 (Table~\ref{tab:results}): dry coughs, the most common presentation, are detected reliably (F1 0.89) and wet coughs at moderate accuracy (F1 0.55). The rarer barking and whooping subtypes are harder (Table~\ref{tab:subtype}): both are acoustically subtler and far less frequent in deployment, so the limited data available for them makes both reliable recognition and reliable evaluation difficult. We therefore treat wet/dry as the dependable operating point today and target the rarer subtypes through ongoing in-domain data collection.

\begin{table}[t]
\centering
\footnotesize
\setlength{\tabcolsep}{5pt}
\begin{tabular}{lcc}
\toprule
\textbf{System} & \textbf{Cough F1} & \textbf{Latency} \\
\midrule
BEATs$^\dagger$          & 59.1\% & 180ms \\
PANNs$^\ddagger$         & 76.4\% & 210ms \\
CoughVID FT$^\S$         & 79.8\% & 95ms \\
\textbf{\system{}}       & \textbf{93.0\%} & 340ms \\
\midrule
\multicolumn{3}{l}{\textit{Additional \system{} metrics}} \\
\quad 3-way Macro-F1     & 0.91 & \\
\quad Wet/Dry Weighted-F1 & 0.75 & \\
\quad Duration Bucket Acc.& 0.91 & \\
\quad Latency p95        & & 520ms \\
\bottomrule
\end{tabular}
\caption{Detection comparison and full \system{} results on 847 in-house conversational audio segments. Detection and 3-way metrics are macro-averaged; the wet/dry subtype metric is weighted-F1 conditional on cough presence. Baselines are off-the-shelf and \emph{not} compute- or backbone-matched; the comparison is capability-illustrative rather than a like-for-like ranking (see text). $^\dagger$\citet{chen2023beats}; $^\ddagger$\citet{kong2020panns}; $^\S$\citet{orlandic2021coughvid} fine-tuned.}
\label{tab:results}
\end{table}

\begin{table}[t]
\centering
\footnotesize
\setlength{\tabcolsep}{4pt}
\begin{tabular}{lccccc}
\toprule
\textbf{System} & \rotatebox{90}{\,Streaming\,} & \rotatebox{90}{\,Throat Sep.\,} & \rotatebox{90}{\,Subtype\,} & \rotatebox{90}{\,Duration\,} & \rotatebox{90}{\,Dialogue\,} \\
\midrule
BEATs$^\dagger$      & \cmark & \xmark & \xmark & \xmark & \xmark \\
PANNs$^\ddagger$     & \xmark & \xmark & \xmark & \xmark & \xmark \\
CoughVID FT$^\S$     & \cmark & \xmark & \xmark & \xmark & \xmark \\
Coswara CNN          & \xmark & \xmark & \xmark & \xmark & \xmark \\
\textbf{\system{}}   & \cmark & \cmark & \cmark & \cmark & \cmark \\
\bottomrule
\end{tabular}
\caption{Capability comparison across the five functions required for conversational integration, each defined operationally in Section~\ref{sec:system}: \emph{streaming} (turn-aligned rolling-buffer processing), \emph{throat sep.}\ (three-way state detection), \emph{subtype} (four-way subtype classification), \emph{duration} (start--end estimation), and \emph{dialogue} (dialogue-aware gating, Section~\ref{sec:gating}). \system{} is the only evaluated system providing all five. See Table~\ref{tab:results} for citation keys.}
\label{tab:capabilities}
\end{table}

\begin{table}[t]
\centering
\footnotesize
\setlength{\tabcolsep}{6pt}
\begin{tabular}{lccc}
\toprule
\textbf{Class} & \textbf{P} & \textbf{R} & \textbf{F1} \\
\midrule
Coughing        & 0.99 & 0.76 & 0.86 \\
Throat clearing & 0.97 & 0.91 & 0.94 \\
None            & 0.86 & 0.99 & 0.92 \\
\midrule
Accuracy        & \multicolumn{3}{c}{0.92} \\
Macro-F1        & \multicolumn{3}{c}{0.91} \\
\bottomrule
\end{tabular}
\caption{External validation on the AMI Meeting Corpus \citep{carletta2006ami} ($n{=}719$ segments with speech present), using \mllm{} with no domain adaptation: three-way state detection (coughing / throat clearing / none).}
\label{tab:ami}
\end{table}

\begin{table}[t]
\centering
\footnotesize
\setlength{\tabcolsep}{10pt}
\begin{tabular}{lc}
\toprule
\textbf{Subtype} & \textbf{F1} \\
\midrule
Dry      & 0.89 \\
Wet      & 0.55 \\
Barking  & 0.23 \\
Whooping & 0.24 \\
\bottomrule
\end{tabular}
\caption{Per-class F1 for four-way cough subtype classification (conditional on cough presence). The clinically central wet/dry distinction reaches 0.75 weighted-F1 (Table~\ref{tab:results}), with reliable dry detection; barking and whooping are harder and limited by data availability.}
\label{tab:subtype}
\end{table}

\paragraph{User Study.}
As a formative feasibility study, three licensed, practicing U.S.\ registered nurses each interacted with \system{} across 6 structured telehealth scenarios, assessing three dialogue quality dimensions: \textit{response appropriateness} (did the agent correctly branch or escalate?), \textit{conversational naturalness} (was the phrasing contextually appropriate?), and \textit{dialogue efficiency} (did cough signals reduce the need for explicit patient questioning?). Evaluators confirmed appropriate branching behavior in scenarios designed to elicit escalation: the agent correctly altered its dialogue strategy based on detected respiratory state. Check-in phrasing was rated natural and contextually appropriate when triggered correctly. Unprompted subtype identification reduced conversational burden by surfacing respiratory status without requiring direct patient self-report, improving information-gathering efficiency within the dialogue flow. Areas for improvement included detection consistency when coughs co-occur with agent speech and gating behavior during prolonged episodes, directly informing ongoing refinements to the mechanisms in Section~\ref{sec:gating}.

\section{Conclusion}

We presented \system{}, a streaming cough analysis pipeline providing fine-grained respiratory analytics for real-time conversational agents. By combining MLLM-based audio understanding with dialogue-aware gating, \system{} enables conversational systems to reason over paralinguistic health signals, a capability that, to the best of our knowledge, is absent from all prior work. Evaluation demonstrates strong detection accuracy, clinically relevant subtype classification, and sub-500ms latency compatible with natural dialogue flow. Future work includes multilingual validation, expanded paralinguistic signal coverage (wheezing, labored breathing), and larger-scale user studies measuring dialogue success across diverse clinical populations.

\section*{Ethical Considerations}

\paragraph{Privacy.}
Audio processed by \system{} may contain sensitive health information. In deployment, audio is processed transiently without persistent storage, and no content is retained beyond the active session. 

\paragraph{Clinical Limitations.}
\system{} provides informational cough analytics, \textbf{not medical diagnosis}. The system triggers follow-up questions rather than clinical recommendations, and users must be clearly informed of these limitations. The system is designed to augment, not replace, clinical judgment.

\paragraph{Bias and Fairness.}
Cough acoustic characteristics vary across demographics, health conditions, and recording equipment. We acknowledge potential performance disparities and recommend ongoing monitoring across user populations. The MLLM backend's training data composition may introduce biases affecting subtype classification for underrepresented groups.


\section*{Limitations}

Subtype labels involve inherent clinical judgment; ground truth is imperfect, as reflected in moderate inter-annotator agreement ($\kappa$=0.67). Performance may degrade in high-noise environments (SNR $<$5dB) or with heavily compressed codecs not represented in the evaluation set. Very soft coughs or those co-occurring with loud speech may be missed. The current evaluation is English-only; generalization to other languages and accents requires dedicated validation. \system{} targets server-based deployment, so end-to-end latency is bound by server compute rather than on-device resources; the MLLM backend introduces cost and availability constraints, and we do not target edge or fully on-device operation. Our user study ($n$=3) provides initial validation across response appropriateness, naturalness, and efficiency dimensions; larger-scale studies with diverse populations would further strengthen the evidence base.

\bibliography{references}

@article{porter2019diagnostic,
author = {Porter, Paul and Abeyratne, Udantha and Swarnkar, Vinayak and Tan, Jamie and Ng, Ti-wan and Brisbane, Joanna and Speldewinde, Deirdre and Choveaux, Jennifer and Sharan, Roneel and Kosasih, Keegan and Della, Phillip},
year = {2019},
month = {06},
pages = {81},
title = {A prospective multicentre study testing the diagnostic accuracy of an automated cough sound centred analytic system for the identification of common respiratory disorders in children},
volume = {20},
journal = {Respiratory Research},
doi = {10.1186/s12931-019-1046-6}
}

@article{imran2020ai4covid,
    title={AI4COVID-19: AI enabled preliminary diagnosis for COVID-19 from cough samples via an app},
    author={Imran, Ali and Posokhova, Iryna and Qureshi, Haneya N and Masood, Usama and Riaz, Muhammad Sajid and Ali, Kamran and John, Charles N and Hussain, Md Iftikhar and Nabeel, Muhammad},
    journal={Informatics in Medicine Unlocked},
    volume={20},
    pages={100378},
    year={2020},
    publisher={Elsevier},
    doi={10.1016/j.imu.2020.100378}
}

@article{orlandic2021coughvid,
    title={The COUGHVID crowdsourcing dataset, a corpus for the study of large-scale cough analysis algorithms},
    author={Orlandic, Lara and Teijeiro, Tomas and Atienza, David},
    journal={Scientific Data},
    volume={8},
    number={1},
    pages={156},
    year={2021},
    publisher={Nature Publishing Group},
    doi={10.1038/s41597-021-00937-4}
}

@inproceedings{sharma2020coswara,
  title     = {{Coswara — A Database of Breathing, Cough, and Voice Sounds for COVID-19 Diagnosis}},
  author    = {Neeraj Sharma and Prashant Krishnan and Rohit Kumar and Shreyas Ramoji and Srikanth Raj Chetupalli and Nirmala R. and Prasanta Kumar Ghosh and Sriram Ganapathy},
  year      = {2020},
  booktitle = {{Interspeech 2020}},
  pages     = {4811--4815},
  doi       = {10.21437/Interspeech.2020-2768},
  issn      = {2958-1796},
}

@article{drugman2013objective,
  title={Objective Study of Sensor Relevance for Automatic Cough Detection},
  author={Thomas Drugman and J{\'e}r{\^o}me Urbain and Nathalie Bauwens and Ricardo Chessini and Carlos Valderrama and Patrick Lebecque and Thierry Dutoit},
  journal={IEEE Journal of Biomedical and Health Informatics},
  year={2013},
  volume={17},
  pages={699-707},
  url={https://api.semanticscholar.org/CorpusID:14919774}
}

@article{morice2007ers,
    title={{ERS} guidelines on the assessment of cough},
    author={Morice, Alyn H and Fontana, Giovanni A and Belvisi, Maria G and Birring, Surinder S and Chung, Kian Fan and Dicpinigaitis, Peter V and Kastelik, Jacek A and McGarvey, Lorcan P A and Smith, Jaclyn A and Tatar, Milos and Widdicombe, John},
    journal={European Respiratory Journal},
    volume={29},
    number={6},
    pages={1256--1276},
    year={2007},
    publisher={European Respiratory Society},
    doi={10.1183/09031936.00101006}
}

@article{chung2009semantics,
    title={Semantics and types of cough},
    author={Chung, Kian Fan and Bolser, Don and Davenport, Paul and Fontana, Giovanni and Morice, Alyn and Widdicombe, John},
    journal={Pulmonary Pharmacology \& Therapeutics},
    volume={22},
    number={2},
    pages={139--142},
    year={2009},
    publisher={Elsevier},
    doi={10.1016/j.pupt.2008.12.008}
}

@article{laguarta2020covid,
    title={COVID-19 artificial intelligence diagnosis using only cough recordings},
    author={Laguarta, Jordi and Hueto, Ferran and Subirana, Brian},
    journal={IEEE Open Journal of Engineering in Medicine and Biology},
    volume={1},
    pages={275--281},
    year={2020},
    publisher={IEEE},
    doi={10.1109/OJEMB.2020.3026928},
}

@article{kong2020panns,
author = {Kong, Qiuqiang and Cao, Yin and Iqbal, Turab and Wang, Yuxuan and Wang, Wenwu and Plumbley, Mark D.},
title = {PANNs: Large-Scale Pretrained Audio Neural Networks for Audio Pattern Recognition},
year = {2020},
issue_date = {2020},
publisher = {IEEE Press},
volume = {28},
issn = {2329-9290},
url = {https://doi.org/10.1109/TASLP.2020.3030497},
doi = {10.1109/TASLP.2020.3030497},
journal = {IEEE/ACM Trans. Audio, Speech and Lang. Proc.},
month = nov,
pages = {2880–2894},
numpages = {15}
}

@inproceedings{gong2021ast,
    title={AST: Audio Spectrogram Transformer},
    author={Gong, Yuan and Chung, Yu-An and Glass, James},
    booktitle={Proceedings of INTERSPEECH},
    pages={571--575},
    year={2021},
    doi={10.21437/Interspeech.2021-698}
}

@inproceedings{chen2023beats,
author = {Chen, Sanyuan and Wu, Yu and Wang, Chengyi and Liu, Shujie and Tompkins, Daniel and Chen, Zhuo and Che, Wanxiang and Yu, Xiangzhan and Wei, Furu},
title = {BEATs: audio pre-training with acoustic tokenizers},
year = {2023},
publisher = {JMLR.org},
booktitle = {Proceedings of the 40th International Conference on Machine Learning},
articleno = {203},
numpages = {16},
location = {Honolulu, Hawaii, USA},
series = {ICML'23}
}

@inproceedings{gong2023whisper,
  author    = {Gong, Yuan and Khurana, Sameer and Karlinsky, Leonid and Glass, James},
  title     = {Whisper-AT: Noise-Robust Automatic Speech Recognizers are Also Strong General Audio Event Taggers},
  booktitle = {{Interspeech 2023}},
  pages     = {2798--2802},
  year      = {2023},
  doi       = {10.21437/Interspeech.2023-2193},
  issn      = {2958-1796},
}

@article{chu2023qwen,
    title={Qwen-Audio: Advancing universal audio understanding via unified large-scale audio-language models},
    author={Chu, Yunfei and Xu, Jin and Zhou, Xiaohuan and Yang, Qian and Zhang, Shiliang and Yan, Zhijie and Zhou, Chang and Zhou, Jingren},
    journal={arXiv preprint arXiv:2311.07919},
    year={2023}
}

@article{qwen3omni,
    title={Qwen3-Omni Technical Report},
    author={Xu, Jin and Guo, Zhifang and Hu, Hangrui and Chu, Yunfei and Wang, Xiong and He, Jinzheng and Wang, Yuxuan and Shi, Xian and He, Ting and Zhu, Xinfa and others},
    journal={arXiv preprint arXiv:2509.17765},
    year={2025}
}

@inproceedings{litman2014sigdial,
    title={Evaluating a Spoken Dialogue System that Detects and Adapts to User Affective States},
    author={Litman, Diane J. and Forbes-Riley, Kate},
    booktitle={Proceedings of the 15th Annual Meeting of the Special Interest Group on Discourse and Dialogue (SIGDIAL)},
    year={2014},
    url={https://aclanthology.org/W14-4324/},
    organization={Association for Computational Linguistics}
}

@inproceedings{katada2020sigdial,
    title={Is She Truly Enjoying the Conversation?: Analysis of Physiological Signals toward Adaptive Dialogue Systems},
    author={Katada, Shun and Okada, Shogo and Hirano, Yuki and Komatani, Kazunori},
    booktitle={Proceedings of the 2020 International Conference on Multimodal Interaction (ICMI)},
    year={2020},
    doi={10.1145/3382507.3418844},
    organization={ACM}
}

@inproceedings{obi2024sigdial,
    title={Using Respiration for Enhancing Human-Robot Dialogue},
    author={Obi, Chukwuemeka and Funakoshi, Kotaro},
    booktitle={Proceedings of the 25th Annual Meeting of the Special Interest Group on Discourse and Dialogue (SIGDIAL)},
    year={2024},
    organization={Association for Computational Linguistics}
}

@inproceedings{zhao2024affect,
    title={Affect Recognition in Conversations Using Large Language Models},
    author={Feng, Shutong and Sun, Guangzhi and Lubis, Nurul and Wu, Wen and Zhang, Chao and Gasic, Milica},
    booktitle={Proceedings of the 25th Annual Meeting of the Special Interest Group on Discourse and Dialogue (SIGDIAL)},
    pages={259--273},
    year={2024},
    url={https://aclanthology.org/2024.sigdial-1.23/},
    organization={Association for Computational Linguistics}
}

@inproceedings{carletta2006ami,
    title={The {AMI} Meeting Corpus: A Pre-announcement},
    author={Carletta, Jean and Ashby, Simone and Bourban, Sebastien and Flynn, Mike and Guillemot, Mael and Hain, Thomas and Kadlec, Jaroslav and Karaiskos, Vasilis and Kraaij, Wessel and Kronenthal, Melissa and Lathoud, Guillaume and Lincoln, Mike and Lisowska, Agnes and McCowan, Iain and Post, Wilfried and Reidsma, Dennis and Wellner, Pierre},
    booktitle={Machine Learning for Multimodal Interaction (MLMI)},
    pages={28--39},
    year={2006},
    publisher={Springer},
    doi={10.1007/11677482_3}
}

\end{document}